\documentclass[letterpaper, 10 pt, conference]{ieeeconf}  

\IEEEoverridecommandlockouts                              

\usepackage{graphicx} 
\usepackage{booktabs}
\usepackage{xcolor}
\usepackage{float}

\title{\LARGE \bf
Near Real-Time Position Tracking for Robot-Guided Evacuation
}

\author{Mollik Nayyar$^{1}$ and Alan R. Wagner$^{2}$
\thanks{$^{1}$Mollik Nayyar is with the Department of Aerospace Engineering,
        The Pennsylvania State University, University Park, PA, USA
        {\tt\small mollik.nayyar@gmail.com}}%
\thanks{$^{2}$Alan R. Wagner is with the Department of Aerospace Engineering at The Pennsylvania State University
        {\tt\small alan.r.wagner@psu.edu}}%
}

\begin{document}

\maketitle
\thispagestyle{empty}
\pagestyle{empty}

\begin{abstract}


During the evacuation of a building, the rapid and accurate tracking of human evacuees can be used by a guide robot to increase the effectiveness of the evacuation \cite{NayyarWagner2019,nayyar2023learning}. This paper introduces a near real-time human position tracking solution tailored for evacuation robots.  Using a pose detector, our system first identifies human joints in the camera frame in near real-time and then translates the position of these pixels into real-world coordinates via a simple calibration process. We run multiple trials of the system in action in an indoor lab environment and show that the system can achieve an accuracy of 0.55 meters when compared to ground truth. The system can also achieve an average of 3 frames per second (FPS) which was sufficient for our study on robot-guided human evacuation. The potential of our approach extends beyond mere tracking, paving the way for evacuee motion prediction, allowing the robot to proactively respond to human movements during an evacuation.

\end{abstract}

\section{INTRODUCTION}


There are many factors that influence how people evacuate. Debris or lack of visibility may hinder their ability to move to an exit. Injuries or disabilities may prevent them from using certain exits. And disorientation or confusion may increase the hesitancy to evacuate. The most common problem with evacuees is simply that they do not evacuate when an alarm sounds \cite{VANDERWAL2021105121}. Hesitancy to evacuate may prove fatal because during a real emergency the time to reach safety may be limited and existing escape routes may become congested. It has also been observed that the onset of an emergency typically causes uncertainty in the people nearby \cite{leach1994survival, kuligowski2013predicting}. Depending on the type of emergency, people may not evacuate at all, they may freeze and remain motionless, or become compliant blindly following any instructions they encounter. During an evacuation, the behavior of other evacuees nearby is often a determining factor impacting when and how quickly a person evacuates \cite{kuligowski2013predicting, NayyarWagner2019}.  Research using video from real emergency evacuations has shown, however, that having a guide during an evacuation significantly reduces the delay people take prior to evacuating \cite{VANDERWAL2021105121,shields2000study,samochine2005investigation}.

We believe that a robot that acts as a guide can reduce the time required to evacuate and thereby save lives. For a robot to guide people to an exit, it must be able to track the person in near real-time. We seek to use robots to guide evacuees to uncongested exits during an evacuation. A successful evacuation robot will need to keep pace with an evacuee or evacuees and will need to be able to track the position of the people in its local environment. Such tracking will need to be near real-time for the robot to make timely navigation decisions and for it to be able to discern if an evacuee has stopped following it. 

If robots are going to serve as evacuation guides, then they will likely need to use the existing available camera infrastructure. For buildings such as schools and high-rise residences, locations that might be best served by an evacuation robot, the potential cost of adding a new camera or motion tracking system would inhibit the use of evacuation robots. On the other hand, many buildings have an existing security or surveillance camera system. If a standard resolution security camera system could be used to provide perceptual information, then the adoption of robots for evacuation is more affordable and therefore more likely. For this reason, our work has focused on the development of perceptual techniques that could use the existing camera infrastructure to provide 1-3Hz centralized evacuee position information. If such a system could be developed, then the need for expensive, fast perceptual processing on the robot would be reduced.

Our work thus focuses on a computationally efficient, vision-based tracking system, that uses static cameras located indoors and can be realized using off-the-shelf components. The methodology presented here relies on open-source pose detection models and a simple calibration process that does not need camera intrinsic matrix or distortion model. The system can work with low resolution cameras that may already be available in many buildings and does not require large datasets to be collected for any new environment. A small set of calibration images are sufficient to generate the camera to world space model. The lower accuracy of our system compared to traditional motion capture systems is compensated for by its simplicity, affordability, and general applicability. In the next section, we first discuss the position estimation system and the methodology used for generating position estimates. Then we present the modifications needed for making it near real-time and finally, we present results and conclusions on the accuracy of the system operating in a new environment.

\section{Indoor Position Estimation System}
Before we discuss the near real-time system, we first present a general camera-based position estimation system. This system was originally developed for a physical robot-guided evacuation experiment \cite{nayyar2023learning}. The experiment consisted of running a total of 106 subjects in individual and group conditions with a robot available for guidance during an emergency. The subjects were asked to perform a reading task and were not informed about the emergency. During the task, three fire alarms placed in the environment were activated unbeknownst to the evacuees. The objective was to observe and collect data on evacuee behavior during a robot-guided evacuation. Evacuee motion data was then used to create a model of evacuee behavior during the evacuation. A system of four cameras were used to cover the entire space and videos of evacuee motion were collected for post processing and model creation. The following subsections discuss the pose detection and camera calibration steps in more detail. 


\subsection{Human Pose Detection}
\begin{figure} [thb]
\centering
\includegraphics[width=2.5in]{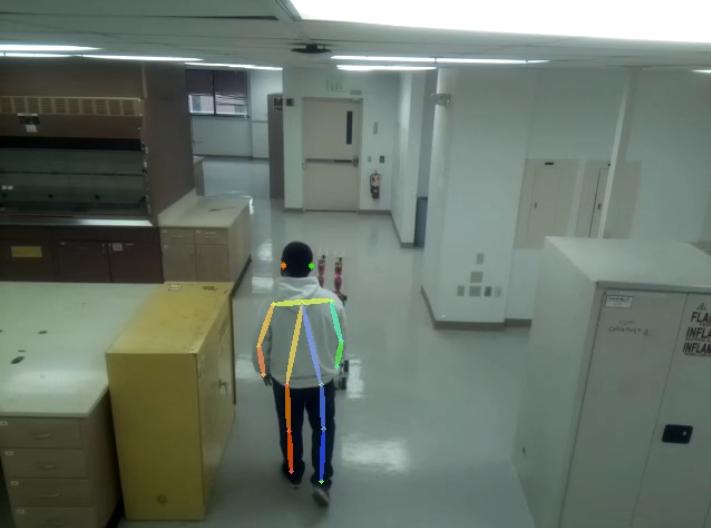}
\caption{The figure shows pose output from the AlphaPose model of a single subject following the robot. The different keypoints representing the body locations can be seen. The ankle keypoints were used for estimating the subject's position in the environment.}
\label{fig:PoseDetec}
\end{figure}

Human poses in the environment were extracted using an open source deep learning library called AlphaPose \cite{AlphaPosefang2017rmpe}. The specific AlphaPose model version used was the YoloV3 model \cite{redmon2018yolov3} with a ResNet152 backbone trained on the COCO Keypoint 2017 dataset with 17 body keypoints. This model was used for evacuee pose detection on 640x480 resolution videos collected during the experiment. The left ankle keypoint was used to determine the location of the subject in the environment. Figure \ref{fig:PoseDetec} shows an example of the pose detections of one of the subjects during the experiment. A different YoloV3 model was trained to detect the robot in the environment to extract the bounding box locations of the robot. The bounding box was then converted to a pixel point in the camera frame which was then used to estimate the robot's position using the camera model.

\subsection{Camera – World Space conversion}
\begin{figure}[t]
\centering
\includegraphics[width=2.5in]{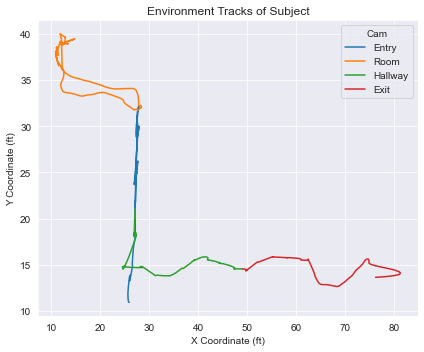}
\caption{The figure shows a global track of a single subject in the shepherding condition. The different colors correspond to the tracks generated from the different cameras placed in the environment.}
\label{fig:EnvironmentTrack}
\end{figure}

\begin{figure} [b]
\centering
\includegraphics[width=2.5in]{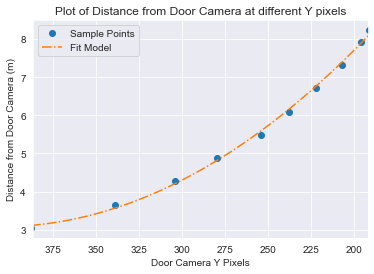}
\caption{The figure shows the pixel to distance mapping for the y-axis of the camera frame.}
\label{fig:camModel}
\end{figure}

To convert the pixels of the detected keypoints to world space, calibration images were collected in the environment at known ground truth distances from the cameras and the AlphaPose model was used to obtain the keypoint locations in pixels. These distance-pixel calibrations were used to create a polynomial model relating the image frame y-axis pixel location to the distance from the camera. For the X-axis (or horizontal distance in the image frame), the width of a calibration object in pixels and inches taken at different y-axis locations was used to create a model for the X direction. Combining the two, we obtained the coordinates of the subject with respect to the camera. This was then transformed into world space coordinates by incorporating the camera’s world space coordinates. This conversion was performed for each of the cameras to obtain a global track of the subject in the environment as shown in Figure \ref{fig:EnvironmentTrack}. An example of the camera model is shown in Figure \ref{fig:camModel}. For some cameras, the positive $X$ and $Y$ axes of the camera had a different alignment with the positive $X$ axis and $Y$ axis in the world space. This change in orientation of the cameras was accounted for before making the relevant coordinate transformations for the subject positions.

\begin{figure}[thb]
\centering
\includegraphics[width=2.5in]{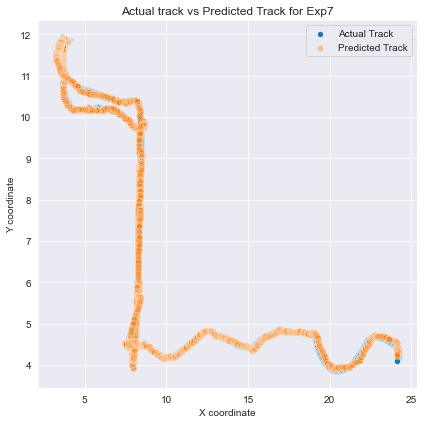}
\caption{The figure shows the actual and the predicted track of one of the subjects during the physical experiment. The distance on the axes are in meters.}
\label{fig:modelPrediction}
\end{figure}

Finally, the tracks of both the evacuee and the robot were used to train an autoregressive motion prediction model that used the past positions of the evacuee and the robot to predict the position of the evacuee 0.25 seconds in the future. The actual and predicted tracks are shown in the Fig. \ref{fig:modelPrediction}. 

\section{Near Real-Time Setup}
For effective robot-guided evacuations the position estimates must be available to the robot in near real-time. One of the limitations of the system discussed thus far was the need for post-processing of the videos. This prompted the need to develop a near real-time system of position estimation. For us, near real-time is approximately 1-2 HZ. Additionally, since it was not possible to obtain ground truth tracks of the subjects during the physical experiment, the accuracy of the position estimates could not be determined. To address these concerns, we decided to set up a similar system in a new environment and incorporate ground truth accuracy measurements in a near real-time position estimation setting. The subsequent sections discuss the near real-time setup used for this work and the results from the trial runs of the system.
There are two essential elements to a real-time position tracking system. First, the ability to acquire camera frames from multiple cameras quickly and second, performing computationally efficient image processing on the frames for detection of humans and extracting position estimates. To ensure that the system does not get overloaded with new frames, older frames may need to be dropped to perform image processing on new frames as soon as they are available. To achieve this, we use real-time acquisition and transmission of frames using the 'imagezmq` image transport library \cite{imagezmq}, a lightweight peer to peer message passing library built on ZMQ and its PyZMQ bindings. Similar to data messages in ROS (Robot operating system), imagezmq is also a message passing protocol but is optimized for speed and ease of use for opencv images. It provides a couple of message passing protocols namely REQ/REP and PUB/SUB.
\begin{itemize}
    \item  In the REQ/REP messaging protocol, each image sender must await a REPLY before continuing. This is a blocking protocol.
    \item In PUB/SUB, each image sender sends an image, but does not expect a REPLY from the central image hub. It can continue sending images without waiting for an acknowledgement from the image hub. This is a non-blocking protocol. 
\end{itemize}
ROS is built on a PUB/SUB protocol by default (it also provides a REQ/REP system using services), however, ROS requires all systems on the network to be able to operate ROS nodes and connect to a singular ROS master. On the other hand, imagezmq operates as python library with minimal dependencies and does not rely on a message broker system like rosmaster and instead uses much more efficient peer to peer system without any additional process overhead. Additionally, it allows setting up multiple camera sources on a local network. 

\subsection{Hardware setup}
Like the system mentioned in the previous section, we used two Raspberry Pis as the client image sender systems with a Raspberry Pi camera as the indoor surveillance camera. The Raspberry Pi's were connected to a local network and were configured to start sending frames to a ground station on a static IP address once the frames start getting captured. The system initializes the cameras, establishes a connection to the ground station and then starts sending the unprocessed camera frames. However, since the system is set up in a REQ/REP protocol, the Raspberry Pis will only send a frame if they receive a reply from the ground station that the previous frame was received. This allows the ground station to perform image processing on an image frame and then send the reply to the cameras for the next one ensuring that processing is performed on the latest frames and does not queue up old frames in a buffer.

\subsection{Pose Estimation}
For this work, a new and light weight model was used for pose detection. Poses were extracted using an open-source deep learning model called YOLOv7 \cite{wang2022YOLOv7}. Typically, Yolo models are well known for their fast object detection but for this work YOLOv7 model's pose estimation pipeline was used. It is trained on the COCO Keypoint 2017 dataset with 17 body keypoints which is consistent with the AlphaPose model used previously.

\section{Result}
\begin{figure} [tbh]
\centering
\includegraphics[width=2.5in]{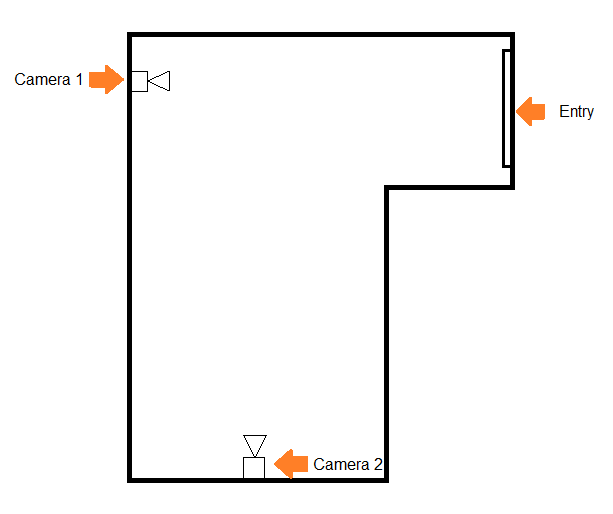}
\caption{The figure shows the layout of the test environment for this work.}
\label{fig:layout}
\end{figure}

\begin{figure} [tbh]
\centering
\includegraphics[width=2.8in]{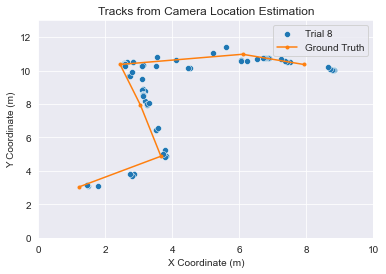}
\caption{The figure shows the track of the subject in trial 8. Position estimates from the camera based position estimation is shown as blue dots. The ground truth track of the subject is shown by the yellow line.}
\label{fig:track}
\end{figure}

The near real-time position tracking system described in this paper was used to track the positions of a subject in the environment. 20 trials of a subject's motion were collected with different motions in each track. Environmental markers were used to extract the ground truth track of the subject in each trial. A grid of 2ft x 2ft cells was marked on the environment floor. Each cell represented the world space coordinates of the center of the cell. The subject moved and stood over position marked cells during the trials. The ground truth positions of the subject's motion were extracted from the cells that the subject stood over during the trials. The Fig. \ref{fig:layout} shows the layout of the environment and the locations of the cameras. The actual track and the estimated positions from the tracking system are presented in the Fig. \ref{fig:track}. Due to noise and erroneous detections in the pose detection model inference step, values that fell beyond 1.5 times the Inter-Quartile Range (IQR) were filtered out. The aggregated mean and standard deviation of the estimated position error across all the 20 trials is (M=0.556, SD=0.069) in meters with a minimum error of 0.43 meters and a maximum error of 0.71 meters. The average FPS across all the 20 trials was (M=3.063, SD=0.238) with a minimum of 2.6 and a maximum of 3.45.
\begin{figure} [tbh]
\centering
\includegraphics[width=2.8in]{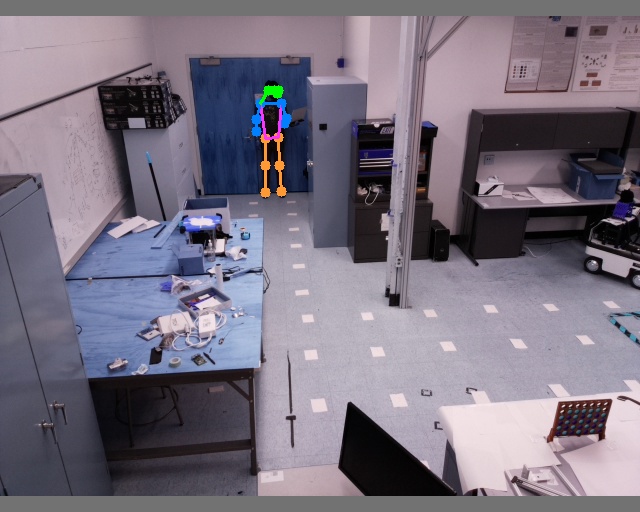}
\caption{The figure shows the subject in the new environment at the entrance (marked 'Entry' in Figure 5). The pose estimates from the Yolov7 model are super imposed on the image. The grid of white squares on the floor are environmental markers with position coordinates.}
\label{fig:exp}
\end{figure}

\section{CONCLUSIONS}
In this paper, we demonstrate a near real-time position tracking system using inexpensive, off-the-shelf components. The results suggest that the system can achieve an accuracy of 0.55 meters with an average frame rate of 3 frames per second with a two-camera setup. Through this system, tracking individuals can become a challenge as small errors in detection and noise make it hard to distinguish the positions of two more closely situated individuals. Such erroneous detections can poison the robot's estimate of the number of people and their positions. Additionally, a faster frame rate may be desired in some applications. Any process that causes the ground station system to lag will create inconsistent position stamps thereby decreasing the frame rate even further. Despite some of these limitations, the system was capable of detecting the location of the humans subject in the environment and we observe in Fig. \ref{fig:track} that detected tracks closely follow the ground truth track. By sequentially tracking the real-time positions of evacuees, the data can be harnessed to create a predictive model such as one used in \cite{nayyar2023learning}. Such models could perhaps forecast potential evacuee trajectories, enabling the robot to not only follow but anticipate human movements. In doing so, the robot can strategize its actions more effectively, ensuring a smoother, safer evacuation process.

\section*{ACKNOWLEDGMENT}
This material is based upon work supported by the National Science Foundation under Grant Number CNS-1830390 and IIS-2045146. Any opinions, findings, and conclusions or recommendations expressed in this material are those of the authors and do not necessarily reflect the views of the National Science Foundation.





\bibliographystyle{IEEEtran}
\bibliography{ref}

\end{document}